\documentclass[11pt]{article}

\usepackage[final]{acl}

\usepackage{float}
\usepackage{times}
\usepackage{latexsym}
\usepackage[T1]{fontenc}
\usepackage[utf8]{inputenc}
\usepackage{microtype}
\usepackage{inconsolata}
\usepackage{graphicx}
\usepackage{booktabs}
\usepackage{tabularx}
\usepackage{amsmath}
\usepackage{multirow}
\usepackage{xcolor}
\usepackage{url}
\usepackage{placeins}

\newcommand{\same}[1]{{\color{gray}\fontsize{7pt}{7pt}\selectfont\hspace{1pt}(\pm0.000)}}

\usepackage[most]{tcolorbox}
\usepackage{listings}
\usepackage{xcolor}

\newtcblisting{PromptBlock}[2][gray]{%
  enhanced,
  breakable,
  listing only,
  colback=#1!3,
  colframe=#1!55!black,
  coltitle=white,
  title={#2},
  fonttitle=\bfseries\small,
  listing options={%
    basicstyle=\ttfamily\scriptsize,
    breaklines=true,
    breakatwhitespace=true,
    columns=fullflexible,
    keepspaces=true,
    showstringspaces=false,
    upquote=true
  },
  left=1.2mm,
  right=1.2mm,
  top=1mm,
  bottom=1mm,
  boxsep=1mm,
  before skip=6pt,
  after skip=6pt
}

\title{Rubric Semantics or Numerical Priors? Auditing Scale-Direction Bias in LLM Judges}

\author{
  Chanwoo Kim, Doyun Kim, Minsoo Song, Sihyeon Lee, Chanjun Park\textsuperscript{$\dagger$} \\
  \normalfont Soongsil University \\
  {\small\texttt{\{kcw4776, doyunkim, ecoses042, sihyeon2988\}@soongsil.ac.kr}} \\
  {\small\texttt{chanjun.park@ssu.ac.kr}}
}
\begin{document}
\maketitle
\begingroup
\renewcommand{\thefootnote}{\ensuremath{\dagger}}
\footnotetext{Corresponding author.}
\endgroup

\begin{abstract}
Recent benchmark evaluations increasingly rely on Large Language Models (LLMs) as automatic judges. This paradigm assumes LLMs evaluate based on the rubric's semantic meaning rather than its numeric scale. We question this assumption by introducing \textit{semantic faithfulness}: the principle that semantically equivalent rubrics must yield consistent scores. To quantify this, we propose a scale-reversal probe that inverts score anchors without altering evaluation criteria. Furthermore, we introduce \textit{floor normalization} to calibrate reversal disagreement against intrinsic model variability, enabling fair cross-model comparisons. Experiments across five meta-evaluation benchmarks with seven diverse LLM judges spanning open-weight and proprietary models reveal widespread violations of semantic faithfulness, showing that raw reversal agreement alone is misleading. Finally, we demonstrate that bidirectional anchor averaging reduces dependence on any single anchor direction without compromising alignment with human judgments. Our findings emphasize that trustworthy LLM judges must reliably follow rubric semantics, not just correlate with human ratings.
\end{abstract}

\begin{figure}[t]
  \centering
  \includegraphics[width=\columnwidth]{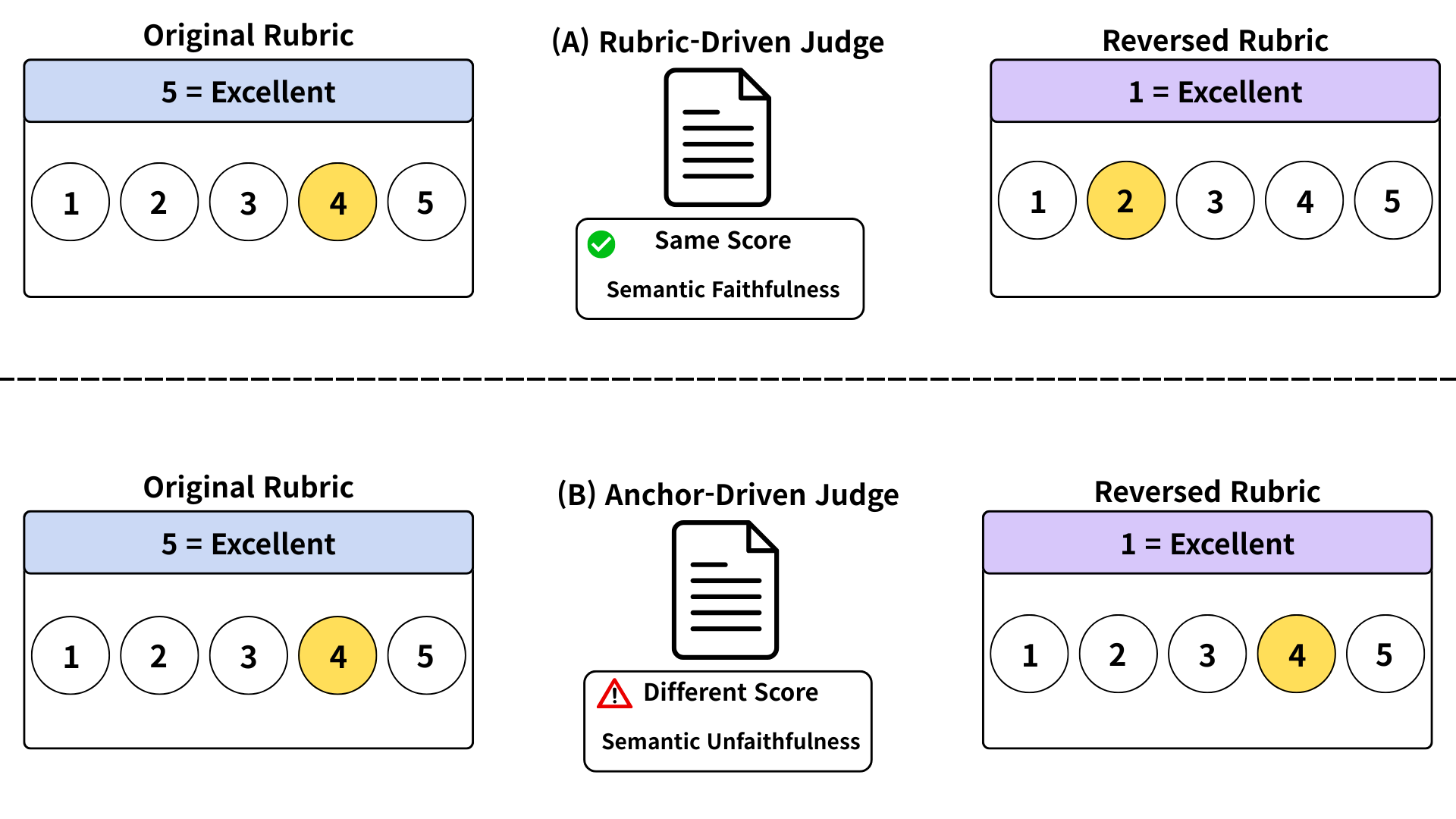}
  \caption{Semantic faithfulness under scale reversal (schematic). The two rubrics are identical except for their numerical encoding: ``5 = Excellent'' becomes ``1 = Excellent''. (A) A rubric-driven judge follows the anchor semantics and flips the number it emits, giving the same score after remapping. (B) An anchor-driven judge keeps the same number regardless of what the anchor means, so the same response is graded differently depending only on score direction.}
  \label{fig:experiments}
\end{figure}
\section{Introduction}
Large language models (LLMs) have emerged as the prevailing standard for evaluating open-ended text generation. Unlike traditional reference-based metrics, such as ROUGE and BERTScore \citep{lin-2004-rouge,zhang2019bertscore,sellam2020bleurt}, which require often-unavailable gold references and struggle to capture nuanced quality, the \emph{LLM-as-a-Judge} paradigm scales evaluation by prompting models to assign ordinal scores based on qualitative rubrics. 

Despite this utility, LLM evaluators suffer from well-documented vulnerabilities, including position and verbosity biases \citep{zheng2023judging}. Furthermore, while existing meta-evaluations predominantly focus on alignment with human preferences, they often overlook a foundational assumption of rubric-based assessment: numerical labels should function strictly as arbitrary indices for qualitative descriptors. A robust judge must evaluate the underlying criteria consistently, irrespective of how the numerical scale is mapped. To systematically evaluate whether an LLM judge exhibits semantic faithfulness, which we operationalize here through the criterion of numerical encoding invariance, we introduce scale reversal as a diagnostic probe and quantify the resulting bias using the floor-normalized reversal deficit. If a model fails to maintain this invariance, it exposes a superficial reliance on numerical anchors rather than a true comprehension of the rubric, fundamentally undermining the validity of the evaluation.

Under our scale reversal probe, we invert the standard numerical mapping, such as assigning a score of one to the highest quality instead of five, while preserving the textual criteria. An invariant judge adapts its output to the newly assigned integer, yielding an identical evaluation upon remapping. However, raw disagreement under scale reversal is an insufficient metric because LLM judges exhibit intrinsic decoding variance, even under deterministic settings. Raw reversal agreement is often entangled with a model's baseline self-reproduction capability, which can incorrectly rank judge reliability. The floor-normalized reversal deficit resolves this by calibrating the reversal discrepancy against the model's test-retest consistency, effectively isolating genuine encoding bias from inherent generation noise.

We apply this framework to audit seven LLM judges across five meta-evaluation benchmarks. Furthermore, we introduce bidirectional anchor averaging to reduce reliance on choosing a single scale direction and evaluate its alignment with human judgments and ground-truth labels. In summary, our contributions are threefold. First, we formalize the taxonomy of semantic faithfulness via numerical encoding invariance, measured by the floor-normalized reversal deficit under a controlled scale reversal probe. Second, we demonstrate that floor normalization is essential to decouple numerical encoding bias from intrinsic decoding variance, correcting the misidentification of robust judges caused by raw agreement metrics. Third, we provide a comprehensive audit identifying where LLMs fail to generalize beyond numerical anchors, and we introduce bidirectional anchor averaging as a complementary inference-time remedy when evaluations can be scored in both directions.

\section{Related Work}
\subsection{LLMs as Automatic Judges}
Large language models have become the dominant paradigm for automatic evaluation across a wide range of NLP tasks. Compared with traditional reference-based metrics such as ROUGE \citep{lin-2004-rouge}, BERTScore \citep{zhang2019bertscore}, and BLEURT \citep{sellam2020bleurt}, LLM judges provide greater flexibility by evaluating outputs directly from natural-language instructions and rubrics \citep{zheng2023judging, gu2024survey}. This paradigm has been widely adopted for benchmarking open-ended generation, instruction following, dialogue, and summarization, where reference-based metrics are often insufficient. LLM judges underpin public leaderboards and benchmarks.

\subsection{Reliability of LLM Judges}
Despite their success, recent work has identified numerous sources of bias in LLM-based evaluation. Existing studies have reported position bias, verbosity bias, self-enhancement bias, and preference instability under different prompt formulations \citep{zheng2023judging,chen2024humans,ye2025justice}. Other work has investigated calibration \citep{liu2024calibrating}, consistency \citep{stureborg2024large,haldar2025rating}, and agreement with human annotators \citep{liu2023g,thakur2025judging}, demonstrating that evaluation quality depends on prompt design, model choice, and inference settings. These studies motivate evaluating both agreement with external judgments and sensitivity to scoring-prompt design.

Scoring biases associated with rubric order, score IDs, and reference-answer scores have also been investigated \citep{li2026evaluating}. Latent numerical preferences estimated through random-number generation, including task-conditioned prompts, have been used to calibrate score-token probabilities \citep{asato2026mitigating}. Our study complements these approaches by reversing the mapping between numerical scores and semantic anchors, remapping the outputs to a common scale, and comparing cross-direction disagreement with repeated judgments under an unchanged encoding. This provides a behavioral audit of scale-direction sensitivity relative to repeatability.

\subsection{Meta-Evaluation of Automatic Evaluation}
Meta-evaluation studies assess the reliability of evaluation methods rather than the quality of generated outputs \citep{deutsch2022limitations}. Prior work typically measures agreement with human judgments \citep{mathur2020tangled,math11041006}, inter-annotator consistency, or system-level rank correlation across evaluation metrics \citep{deutsch2022re,deutsch2021statistical}. While these analyses quantify overall evaluation quality, they implicitly assume that the semantics of the evaluation rubric are faithfully followed by the evaluator. In contrast, we introduce semantic faithfulness, which asks whether semantically equivalent rubric formulations lead to equivalent judgments. Our work therefore complements existing meta-evaluation by assessing robustness to semantically invariant changes in rubric encoding against a same-direction repeatability baseline.

\section{Methodology}
\label{sec:method}
\subsection{Semantic Faithfulness via Scale Reversal}

We investigate the \emph{semantic faithfulness} of LLM judges: the extent to which they rely on qualitative rubric descriptions rather than superficial numerical labels. Existing frameworks assume discrete scores merely index semantic categories, implying that rubrics differing only in numeric mapping should yield equivalent evaluations.

To test this assumption, we introduce \textit{Scale Reversal} as a diagnostic probe. Given a standard ordinal scale $\{s_{\min},\dots,s_{\max}\}$ where higher values denote superior quality, we construct a semantically isomorphic rubric by reversing the numeric-to-text alignment. For example, a standard anchor of ``5 = Excellent'' is transposed to ``1 = Excellent'', while all underlying evaluation criteria are strictly preserved.

Let $s^{\mathrm{n}}$ and $s^{\mathrm{r}}$ be the scores predicted under the standard and reversed rubrics, respectively. To make them strictly comparable, we project the reversed score back to the standard scale to obtain a remapped score $\tilde{s}^{\mathrm{r}}$, as defined in Equation~\ref{eq:remap}:
\begin{equation}
\tilde{s}^{\mathrm{r}} = (s_{\min} + s_{\max}) - s^{\mathrm{r}}
\label{eq:remap}
\end{equation}
For a standard 5-point scale, Equation~\ref{eq:remap} simplifies to $6 - s^{\mathrm{r}}$. 

If an LLM judge follows rubric semantics, reversal should change the number it emits but not the judgment behind that number: a response the judge considers excellent receives the top anchor under either rubric, which is $s_{\max}$ on the standard scale and $s_{\min}$ on the reversed one. Systematic disagreement indicates that the judge is influenced by the numerical representation of the rubric, revealing \emph{scale-direction bias} rather than purely semantic reasoning.

\subsection{Floor Normalization}

Observed disagreements under scale reversal cannot be exclusively attributed to semantic bias, as LLM judges inherently demonstrate variance even with deterministic decoding. An evaluator lacking test--retest reliability will naturally yield lower agreement rates, irrespective of the scale's orientation.

Rather than attempting a causal decomposition of semantic and stochastic effects, the deficit should be interpreted as excess reversal sensitivity relative to repeated normal evaluation. To quantify this, we propose \textit{floor normalization}. Specifically, we conduct a secondary evaluation using the standard rubric (\texttt{normal2}) and define two exact-match agreement metrics across the evaluated instances:

\begin{itemize}
    \item \textbf{Flip agreement} ($A_{\mathrm{flip}}$): The proportion of instances where the standard prediction matches the remapped reversed prediction.
    \item \textbf{Floor agreement} ($A_{\mathrm{floor}}$): The baseline consistency, representing the agreement rate between two independent runs under the identical standard rubric.
\end{itemize}

To quantify the semantic bias that persists beyond the evaluator's inherent instability, we compute the difference between these two agreement rates, denoted as $\Delta_{\mathrm{exact}}$ in Equation~\ref{eq:delta_exact}:

\begin{equation}
\Delta_{\mathrm{exact}} = A_{\mathrm{floor}} - A_{\mathrm{flip}}.
\label{eq:delta_exact}
\end{equation}

As formulated in Equation~\ref{eq:delta_exact}, $\Delta_{\mathrm{exact}}$ isolates the impact of score directionality; a larger value signifies a greater sensitivity to the scale orientation. Consequently, we determine that a divergence reflects authentic scale-direction bias only if $\Delta_{\mathrm{exact}}$ strictly exceeds a calibrated tolerance margin $\varepsilon$ (refer to Appendix~\ref{app:epsilon} for calibration details).

We also assess sensitivity to the choice of repeat baseline by adding a second reversed pass. Appendix~\ref{app:fourpass-conditions} defines the four scoring conditions, and Appendix~\ref{app:reverse-repeat} compares normal-repeat and reversed-repeat agreement with cross-direction agreement on the same observations.

\subsection{Leaderboard Dependence and Bidirectional Anchor Averaging}

Although instance-level semantic bias is evident, its impact on aggregated leaderboards requires formal quantification. To assess sensitivity to anchor direction, Equation~\ref{eq:taudir} defines $\tau_{\mathrm{dir}}$ as the system-level Kendall's $\tau_b$ correlation between rankings derived from standard ($\bar{s}^{\mathrm{n}}$) and remapped reversed ($\bar{\tilde{s}}^{\mathrm{r}}$) mean scores:
\begin{equation}
\tau_{\mathrm{dir}} = \tau\!\left(\bar{s}^{\mathrm{n}},\,\bar{\tilde{s}}^{\mathrm{r}}\right)
\label{eq:taudir}
\end{equation}
A $\tau_{\mathrm{dir}} < 1$ indicates that the leaderboard changes depending on the chosen anchor direction, forcing an arbitrary choice of scale. While bounded by the judge's intrinsic reproducibility, this metric highlights a raw directional disagreement rather than definitive distortion.

To reduce reliance on choosing a single scale direction, we propose \emph{bidirectional anchor averaging}: the mean of the standard score and the reversed score remapped to the standard scale. This inference-time adjustment combines both directions without changing the judge model and requires one additional scoring pass. It does not assume that the errors induced by the two encodings are equal and opposite.

We assess the effect of averaging on alignment with human ground truth using item- and system-level rank correlations ($\tau_{\mathrm{item}}, \tau_{\mathrm{sys}}$) for graded ratings and verdict match rates for pairwise setups. These comparisons evaluate its empirical utility rather than assuming that averaging preserves reliability in every setting.

\section{Experimental Setup}
\subsection{Evaluation Benchmarks}
We implement the reversal probe across five meta-evaluation benchmarks. To confirm that the observed effects are not merely artifacts of specific evaluation formats, we utilize the originally published system outputs and reference labels while diversifying the rubric structures from fixed to instance-specific, as well as varying the scale lengths between five-point and ten-point ranges. We classify these benchmarks into two distinct groups based on their ground-truth formats.

\begin{itemize}
    \item \textbf{Benchmarks with Graded Human Ratings:} \textsc{SummEval} \citep{fabbri2021summeval} evaluates 1,600 summaries for coherence, consistency, and relevance using a fixed five-point scale. We construct our probe for this dataset by inverting only the semantic anchors, such as mapping the highest score to the lowest semantic rating and vice versa. Furthermore, \textsc{BiGGen-Bench} \citep{kim2024biggen} assesses 2,780 responses through instance-specific rubrics. In this case, we apply the reversal process by permuting solely the semantic descriptions associated with each specific score.

    \item \textbf{Benchmarks with Pairwise Ground Truth:} \textsc{MT-Bench} \citep{zheng2023judging} measures 960 multi-turn responses against expert preferences using a ten-point scale. Additionally, \textsc{JudgeBench} \citep{tan2024judgebench} evaluates objective correctness by providing 620 response pairs wherein exactly one response is verifiably true. Finally, \textsc{RewardBench} \citep{lambert2025rewardbench} measures human preference and safety across 800 response pairs, strictly excluding any subsets that overlap with MT-Bench. For all datasets within this category, responses are scored individually prior to deriving a final pairwise verdict.
\end{itemize}

\subsection{Judge Models}

To investigate whether semantic faithfulness generalizes across architectures and scales, we evaluate a diverse suite of instruction-tuned LLM judges encompassing various model families, parameter counts, and accessibility types, including both open-weight models and proprietary APIs.

For the open-weight models, we evaluate Gemma-4-31B-it, Qwen3.5-27B\footnote{https://huggingface.co/Qwen/Qwen3.5-27B}, Qwen3.5-122B-A10B\footnote{https://huggingface.co/Qwen/Qwen3.5-122B-A10B}, and gpt-oss-120B \citep{team2026gemma, agarwal2025gpt} at full scale across all five benchmarks. Spanning three distinct families, these models confirm that the observed effects are intrinsic to the judges rather than data artifacts. To reflect real-world evaluation pipelines, we also include three proprietary API models: gpt-4o-mini, gpt-5-mini and gpt-5.4 \citep{hurst2024gpt, singh2025openai}, the latter being a reasoning model that utilizes inference-time computation.

To minimize stochastic variability, all seven judges employ greedy decoding by setting the temperature to $0$. Open-weight models are served using vLLM \citep{kwon2023efficient}, while API models are queried via standard endpoints. Full prompt templates are provided in Appendix~\ref{app:prompts}.

\subsection{Evaluation Protocol}
Each judge evaluates every example under three conditions:
\begin{itemize}
    \item \textbf{Normal}: evaluation using the original rubric.
    \item \textbf{Reversed}: evaluation using the scale-reversed rubric.
    \item \textbf{Normal2}: a repeated evaluation using the original rubric for estimating intrinsic test--retest consistency.
\end{itemize}

\begin{table*}[t]
\centering
\footnotesize
\setlength{\tabcolsep}{2.2pt}
\scalebox{0.9}{%
\begin{tabular*}{\textwidth}{@{\extracolsep{\fill}}lccccccccccccccc@{}}
\toprule
& \multicolumn{3}{c}{\textit{SummEval}} & \multicolumn{3}{c}{\textit{BiGGen-Bench}} & \multicolumn{3}{c}{\textit{MT-Bench} ($1$--$10$)} & \multicolumn{3}{c}{\textit{JudgeBench}} & \multicolumn{3}{c}{\textit{RewardBench}} \\
\cmidrule(lr){2-4}\cmidrule(lr){5-7}\cmidrule(lr){8-10}\cmidrule(lr){11-13}\cmidrule(lr){14-16}
Judge & $A_{\mathrm{flip}}$ & $A_{\mathrm{floor}}$ & $\Delta$ & $A_{\mathrm{flip}}$ & $A_{\mathrm{floor}}$ & $\Delta$ & $A_{\mathrm{flip}}$ & $A_{\mathrm{floor}}$ & $\Delta$ & $A_{\mathrm{flip}}$ & $A_{\mathrm{floor}}$ & $\Delta$ & $A_{\mathrm{flip}}$ & $A_{\mathrm{floor}}$ & $\Delta$ \\
\midrule
Gemma-4-31B  & 0.808 & 0.977 & 0.169 & 0.847 & 0.970 & 0.123 & 0.623 & 0.956 & \textbf{0.333} & 0.771 & 0.911 & 0.140 & 0.749 & 0.971 & \textbf{0.222} \\
 & \multicolumn{3}{r}{\scriptsize$[+0.157,\,+0.181]$} & \multicolumn{3}{r}{\scriptsize$[+0.108,\,+0.138]$} & \multicolumn{3}{r}{\scriptsize$[+0.293,\,+0.371]$} & \multicolumn{3}{r}{\scriptsize$[+0.116,\,+0.165]$} & \multicolumn{3}{r}{\scriptsize$[+0.201,\,+0.243]$} \\
gpt-oss-120B & 0.705 & 0.782 & 0.077 & 0.788 & 0.832 & 0.044 & 0.575 & 0.699 & 0.124 & 0.757 & 0.779 & 0.022 & 0.726 & 0.807 & 0.082 \\
 & \multicolumn{3}{r}{\scriptsize$[+0.062,\,+0.091]$} & \multicolumn{3}{r}{\scriptsize$[+0.028,\,+0.061]$} & \multicolumn{3}{r}{\scriptsize$[+0.091,\,+0.160]$} & \multicolumn{3}{r}{\scriptsize$[-0.002,\,+0.045]$} & \multicolumn{3}{r}{\scriptsize$[+0.057,\,+0.105]$} \\
Qwen3.5-122B & 0.778 & 0.827 & 0.049 & 0.799 & 0.863 & 0.064 & 0.515 & 0.665 & 0.150 & 0.782 & 0.797 & 0.015 & 0.701 & 0.763 & 0.063 \\
 & \multicolumn{3}{r}{\scriptsize$[+0.037,\,+0.061]$} & \multicolumn{3}{r}{\scriptsize$[+0.048,\,+0.080]$} & \multicolumn{3}{r}{\scriptsize$[+0.115,\,+0.184]$} & \multicolumn{3}{r}{\scriptsize$[-0.012,\,+0.035]$} & \multicolumn{3}{r}{\scriptsize$[+0.039,\,+0.087]$} \\
Qwen3.5-27B  & 0.779 & 0.806 & 0.027 & 0.812 & 0.865 & 0.053 & 0.523 & 0.643 & 0.120 & 0.750 & 0.794 & 0.043 & 0.707 & 0.751 & 0.044 \\
 & \multicolumn{3}{r}{\scriptsize$[+0.013,\,+0.039]$} & \multicolumn{3}{r}{\scriptsize$[+0.037,\,+0.068]$} & \multicolumn{3}{r}{\scriptsize$[+0.085,\,+0.157]$} & \multicolumn{3}{r}{\scriptsize$[+0.018,\,+0.071]$} & \multicolumn{3}{r}{\scriptsize$[+0.020,\,+0.069]$} \\
\midrule
gpt-4o-mini & 0.652 & 0.901 & \textbf{0.249} & 0.700 & 0.942 & \textbf{0.242} & 0.500 & 0.772 & 0.272 & 0.732 & 0.893 & \textbf{0.161} & 0.800 & 0.867 & 0.067 \\
 & \multicolumn{3}{r}{\scriptsize$[+0.211,\,+0.288]$} & \multicolumn{3}{r}{\scriptsize$[+0.142,\,+0.350]$} & \multicolumn{3}{r}{\scriptsize$[+0.194,\,+0.350]$} & \multicolumn{3}{r}{\scriptsize$[+0.054,\,+0.273]$} & \multicolumn{3}{r}{\scriptsize$[-0.033,\,+0.150]$} \\
gpt-5-mini & 0.755 & 0.796 & 0.042 & 0.800 & 0.808 & 0.008 & 0.578 & 0.644 & 0.067 & 0.817 & 0.833 & 0.017 & 0.617 & 0.733 & 0.117 \\
 & \multicolumn{3}{r}{\scriptsize$[+0.022,\,+0.062]$} & \multicolumn{3}{r}{\scriptsize$[-0.075,\,+0.100]$} & \multicolumn{3}{r}{\scriptsize$[-0.017,\,+0.155]$} & \multicolumn{3}{r}{\scriptsize$[-0.100,\,+0.133]$} & \multicolumn{3}{r}{\scriptsize$[+0.000,\,+0.233]$} \\
gpt-5.4 & 0.819 & 0.862 & 0.042 & 0.850 & 0.892 & 0.042 & 0.589 & 0.744 & 0.155 & 0.733 & 0.733 & 0.000 & 0.850 & 0.867 & 0.017 \\
 & \multicolumn{3}{r}{\scriptsize$[+0.024,\,+0.061]$} & \multicolumn{3}{r}{\scriptsize$[-0.025,\,+0.100]$} & \multicolumn{3}{r}{\scriptsize$[+0.072,\,+0.239]$} & \multicolumn{3}{r}{\scriptsize$[-0.117,\,+0.100]$} & \multicolumn{3}{r}{\scriptsize$[-0.100,\,+0.133]$} \\
\bottomrule
\end{tabular*}%
}
\caption{
Item-level semantic faithfulness measured by floor-normalized agreement deficit $\Delta$. Positive values indicate disagreement induced by scale reversal beyond baseline reproducibility. Bracketed values denote 95\% bootstrap intervals based on 2,000 resamples. Boldface indicates the largest deficit per benchmark. Due to varying score scales, where MT-Bench uses a 1 to 10 scale and others use 1 to 5, deficits are only comparable within each benchmark. Bootstrap details and ranking-level deficits are provided in Appendix~\ref{app:bootstrap}.
}
\label{tab:main}
\end{table*}

From these three passes we compute the flip agreement ($A_{\mathrm{flip}}$), floor agreement ($A_{\mathrm{floor}}$) and item-level deficit ($\Delta_{\mathrm{exact}}$) defined in Section~\ref{sec:method}. Unless otherwise specified, all reported statistics are computed over the intersection of instances successfully evaluated under all three conditions.

The supplementary analyses use distinct protocols and complete-case sets. The four-pass repeatability and utility analyses use normal, normal2, reversed, and reversed2, with their own inclusion rules (Appendix~\ref{app:fourpass}). These supplementary observations are not pooled into the original three-pass diagnostic tables.

\section{Results}

Our experiments address four research questions (RQ):

\begin{itemize}
    \item \textbf{RQ1.} Do LLM judges faithfully follow rubric semantics?
    \item \textbf{RQ2.} How can semantic faithfulness be measured reliably?
    \item \textbf{RQ3.} Where do violations of semantic faithfulness concentrate?
    \item \textbf{RQ4.} Can violations of semantic faithfulness be mitigated?
\end{itemize}

\subsection{RQ1: Do LLM Judges Faithfully Follow Rubric Semantics?}

The results provide consistent evidence that current LLM judges are sensitive to the numerical encoding of rubric semantics. We investigate this through scale reversal by renumbering rubric levels while maintaining their semantic definitions. When remapped to the original scale, any disagreement surpassing the judge's inherent decoding noise provides behavioral evidence that numerical encoding influences judgments beyond the rubric's stated semantic mapping. We define a violation as a drop in agreement exceeding a 0.02 consistency threshold. Appendix~\ref{app:epsilon} grounds this threshold in the baseline disagreement observed between two identical passes. Table~\ref{tab:main} reveals non-negative deficits across all 35 judge-benchmark configurations. Strictly positive deficits appear in 34 configurations, with 30 exceeding the threshold. This confirms that standard forward-anchored scores do not purely reflect rubric semantics, which fundamentally challenges their reliability.

Our analysis isolates the reversal effect by using a repeated pass from the same judge as a baseline. This properly subtracts decoding noise, which is an essential correction given that baseline agreements range broadly from 0.643 to 0.977. These deficits are statistically robust, and 25 cases exceed the threshold with confidence intervals strictly above zero. Appendix~\ref{app:bootstrap} outlines the corresponding bootstrap procedure. The remaining cases are mostly concentrated in JudgeBench, which yields the smallest overall deficits. Because exact-match agreement naturally degrades with more score categories, we restrict magnitude comparisons to within individual benchmarks. Furthermore, evaluating MT-Bench on its native ten-point scale verifies that this vulnerability extends beyond standard five-point encodings.

While the severity varies, no tested judge is fully immune. Deficits exhibit significant variance within individual benchmarks. Ranges span from $0.027$ to $0.249$ on SummEval, $0.008$ to $0.242$ on BiGGen-Bench, and $0.017$ to $0.222$ on RewardBench. The deficit ranges for JudgeBench and MT-Bench are $0.000$ to $0.161$ and $0.067$ to $0.333$, respectively. Among open-weight models, Gemma-4-31B consistently produces the largest deficits. Scaling up the model size does not mitigate this issue. The 122-billion parameter Qwen3.5 model demonstrates larger deficits than its 27-billion parameter counterpart on four of the five benchmarks. Proprietary models share this vulnerability, with gpt-4o-mini recording the highest deficit among all judges on SummEval, BiGGen-Bench, and JudgeBench.

\subsection{RQ2: How Can Semantic Faithfulness Be Measured Reliably?}

\begin{figure}[t]
\centering
\includegraphics[width=0.8\columnwidth]{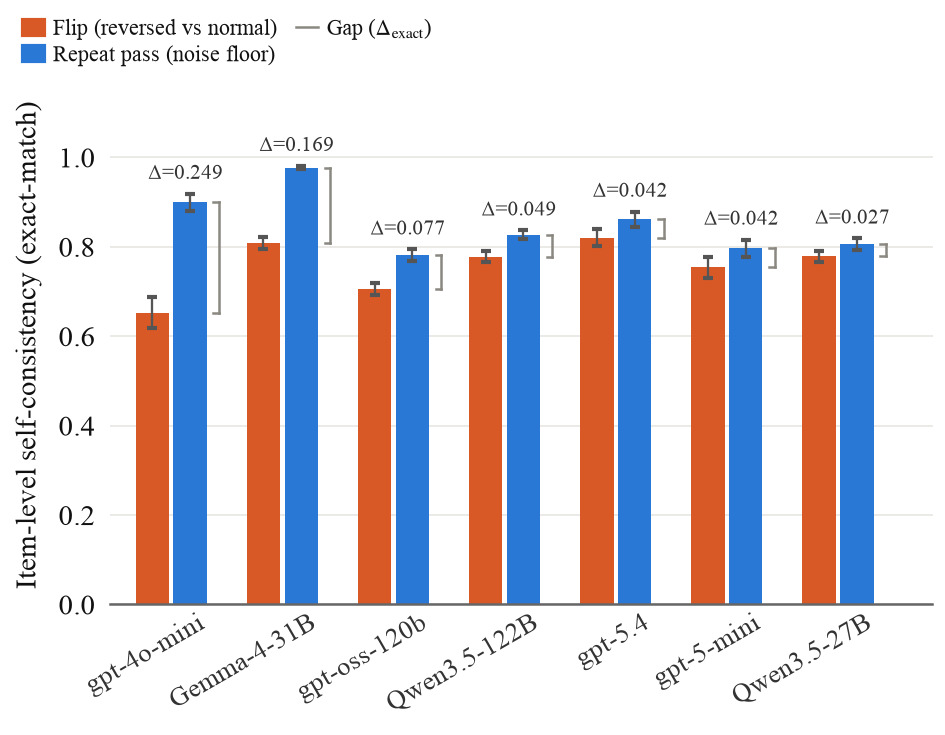}
\caption{
Judge performance on SummEval. $A_{\mathrm{flip}}$ is the agreement between the $\mathrm{normal}$ and reoriented $\mathrm{reversed}$ passes. $A_{\mathrm{floor}}$ is the agreement between the $\mathrm{normal}$ and $\mathrm{normal2}$ passes. The bracket joining the bars indicates the deficit $\Delta_{\mathrm{exact}}$, which is used to sort the judges. Error bars denote $95\%$ cluster bootstrap intervals. While $A_{\mathrm{flip}}$ alone puts Gemma-4-31B and gpt-5.4 within $0.011$ of each other, their deficits differ fourfold when measured against their own floors.
}
\label{fig:floor}
\end{figure}

An intuitive baseline for evaluating semantic faithfulness is the raw agreement between original and reversed assessments, denoted as $A_{\mathrm{flip}}$. However, as illustrated in Figure~\ref{fig:floor}, relying exclusively on this raw metric is insufficient because it is heavily distorted by the inherent reproducibility of the individual judges.

The comparison between Gemma-4-31B and gpt-5.4 on the SummEval benchmark clearly demonstrates this limitation. The raw metric fails to differentiate the two judges, which achieve flip agreements of 0.808 and 0.819, representing the two highest values on this benchmark. Their respective performance floors, however, diverge significantly. Gemma-4-31B replicates its own scores almost perfectly with a floor agreement of 0.977, whereas gpt-5.4 reaches a floor of only 0.862. Consequently, the reversal process incurs a deficit of 0.169 for Gemma-4-31B compared to a mere 0.042 for gpt-5.4. This fourfold difference in performance degradation is entirely obscured by the raw agreement metric. When ranked by $A_{\mathrm{flip}}$, Gemma-4-31B appears as the second most robust of the seven judges, but when evaluated using the floor-normalized deficit, it falls to the sixth position. Furthermore, gpt-oss-120B illustrates this phenomenon from an alternative perspective. It exhibits a substantially lower raw agreement of 0.705 paired with a reduced floor score of 0.782, yielding a deficit of 0.077, which is less than half the magnitude of the deficit observed for Gemma-4-31B.

\begin{figure*}[t]
\centering
\includegraphics[width=0.8\textwidth]{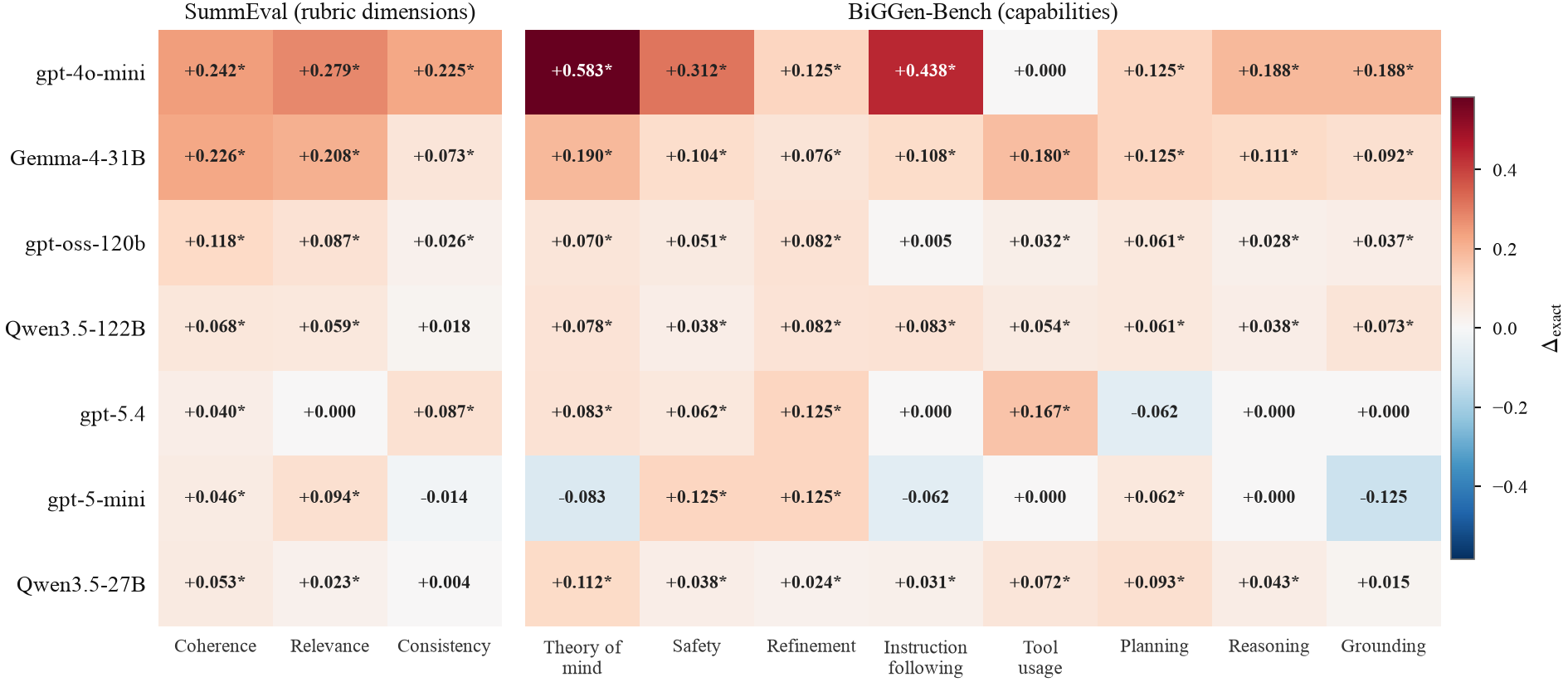}
\caption{
Floor-normalized deficit, denoted as $\Delta_{\mathrm{exact}}$, across judges on SummEval and BiGGen-Bench. We include these two benchmarks because their rubrics isolate specific evaluation criteria, unlike other benchmarks that are partitioned by task domain. Columns are sorted by mean deficit, with asterisks indicating values exceeding $\varepsilon = 0.02$. Both panels share a color scale and a 1 to 5 scoring range, displaying three of the four dimensions for SummEval.
}
\label{fig:dims}
\end{figure*}

This discrepancy extends beyond the SummEval dataset. Across four of the five audited benchmarks, the judge demonstrating the highest raw agreement does not correspond to the judge with the smallest deficit. On MT-Bench, these two ranking orderings invert completely. Specifically, Gemma-4-31B leads in raw agreement with a score of 0.623, yet it simultaneously records the largest deficit among all judges at 0.333. Therefore, interpreting raw agreement as a direct proxy for robustness fundamentally misguides the analytical conclusions across the majority of our evaluations.

These findings collectively establish that semantic faithfulness cannot be reliably measured using raw reversal agreement alone. Rather, the evaluation metric must be explicitly calibrated against the baseline reproducibility of each individual judge. The application of floor normalization fulfills this essential calibration requirement, yielding a deficit measure that facilitates equitable and robust comparisons among judges exhibiting varying degrees of intrinsic consistency.

The reversed-repeat check in Appendix~\ref{app:reverse-repeat} examines whether this diagnostic depends on using only the normal-repeat baseline. On the shared four-pass GPT-4o-mini/SummEval subset, normal-repeat, reversed-repeat, and cross-direction agreement are 90.23\%, 86.62\%, and 65.04\%, respectively. Averaging the two repeat baselines reduces the deficit from 25.19 to 23.39 pp. Thus, excess cross-direction disagreement remains on this subset, although the deficit depends on the chosen repeat baseline and is not a causal decomposition.

\subsection{RQ3: Where Do Violations of Semantic Faithfulness Concentrate?}

Benchmarks with rubrics structured into explicit categories allow us to trace evaluation deficits back to their underlying criteria. Out of the five benchmarks evaluated, only SummEval and BiGGen-Bench exhibit this property. Specifically, SummEval evaluates all instances against fixed dimensions, and BiGGen-Bench employs instance-specific rubrics annotated with targeted capabilities. Consequently, both benchmarks group results based on the specific qualities the judge assesses. We exclude the remaining three benchmarks from this analysis because they utilize a single overarching criterion and categorize items merely by subject matter. Figure~\ref{fig:dims} illustrates the observed deficit for each judge across SummEval and BiGGen-Bench.

\begin{table}[t]
\centering
\small
\begin{tabular*}{\columnwidth}{@{\extracolsep{\fill}}lccc@{}}
\toprule
Judge & $\tau_{\mathrm{dir}}$ & $\tau_{\mathrm{item}}$ & $\Delta\tau_{\mathrm{item}}$ \\
      &                       & {\footnotesize normal $\rightarrow$ avg} & \\
\midrule
\multicolumn{4}{@{}l}{\textit{SummEval}} \\
Gemma-4-31B   & 0.963 & $0.394 \rightarrow 0.408$ & +0.015 \\
gpt-oss-120B  & 0.920 & $0.351 \rightarrow 0.370$ & +0.019 \\
Qwen3.5-122B  & 0.946 & $0.348 \rightarrow 0.369$ & +0.021 \\
Qwen3.5-27B   & 0.912 & $0.352 \rightarrow 0.371$ & +0.019 \\
gpt-4o-mini   & 0.820 & $0.260 \rightarrow 0.270$ & +0.010 \\
gpt-5-mini    & 0.843 & $0.364 \rightarrow 0.382$ & +0.018 \\
gpt-5.4       & 0.837 & $0.331 \rightarrow 0.330$ & $-$0.002 \\
\midrule
\multicolumn{4}{@{}l}{\textit{BiGGen-Bench}} \\
Gemma-4-31B   & 1.000 & $0.380 \rightarrow 0.383$ & +0.003 \\
gpt-oss-120B  & 0.667 & $0.363 \rightarrow 0.377$ & +0.013 \\
Qwen3.5-122B  & 1.000 & $0.384 \rightarrow 0.397$ & +0.013 \\
Qwen3.5-27B   & 1.000 & $0.389 \rightarrow 0.398$ & +0.009 \\
gpt-4o-mini   & 0.286 & $0.316 \rightarrow 0.356$ & +0.040 \\
gpt-5-mini    & 1.000 & $0.338 \rightarrow 0.354$ & +0.016 \\
gpt-5.4       & 1.000 & $0.377 \rightarrow 0.391$ & +0.014 \\
\bottomrule
\end{tabular*}
\caption{
Mitigating violations of semantic faithfulness through bidirectional anchor averaging. $\tau_{\mathrm{dir}}$ measures leaderboard consistency between the two anchor directions, a value below $1$ meaning the leaderboard depends on which direction is used; $\tau_{\mathrm{item}}$ reports agreement with human judgments before and after averaging. This analysis requires human ratings and is therefore run on SummEval and on BiGGen-Bench.
}
\label{tab:remedy}
\end{table}

The two panels demonstrate a largely consistent pattern regarding the fragility of LLM-based judging. On SummEval, six of the seven judges place their largest deficit in coherence or relevance and their smallest in consistency. The sole exception is gpt-5.4, where consistency is instead the most affected dimension at a positive margin of 0.087. Consistency also supplies three of the four cells falling within the $\varepsilon$ threshold, showing a positive 0.018 and a positive 0.004 for the two Qwen judges, and a negative 0.014 for gpt-5-mini. Similarly, on BiGGen-Bench, theory-of-mind ranks among the three most impacted capabilities for six of the seven judges and is the primary source of deficit for three of them. The exception to this trend is gpt-5-mini, where the capability shows a negative margin of 0.083. Conversely, capabilities falling within the $\varepsilon$ margin, such as instruction-following for gpt-oss-120B and grounding for Qwen3.5-27B, are among the least affected.

This pattern suggests that evaluation criteria are more robust when they can be grounded in an external reference. Consistency demonstrates this by verifying whether a summary's claims are entailed by the source text. Grounding, the least affected capability on BiGGen-Bench at a mean positive deviation of 0.040, applies a similar verification for the response as a whole. In contrast, criteria like theory-of-mind, coherence, and relevance require a holistic judgment of the response, which forces the model to dynamically construct its own evaluation standard. We present this view as an interpretative hypothesis rather than a definitive mechanism, given that it explains the extreme cases but fails to account for the entire ranking. For instance, safety is the second most affected capability overall despite resting on comparatively explicit criteria. Additionally, reasoning remains relatively robust despite requiring open-ended inference. Pinpointing the exact causal properties would require rewriting the rubrics while strictly holding the underlying tasks constant.

\begin{table}[t]
\centering
\resizebox{\columnwidth}{!}{%
\begin{tabular}{@{}lcc@{}}
\toprule
Judge & Accuracy & $\Delta$ [95\% CI] \\
      & {\footnotesize normal $\rightarrow$ avg} & \\
\midrule
\multicolumn{3}{@{}l}{\textit{MT-Bench}} \\
Gemma-4-31B   & $0.789 \rightarrow 0.796$ & $+0.007$ [$+0.002$, $+0.014$] \\
gpt-oss-120B  & $0.800 \rightarrow 0.806$ & $+0.007$ [$-0.002$, $+0.016$] \\
Qwen3.5-122B  & $0.796 \rightarrow 0.803$ & $+0.007$ [$-0.001$, $+0.015$] \\
Qwen3.5-27B   & $0.807 \rightarrow 0.813$ & $+0.006$ [$-0.002$, $+0.015$] \\
gpt-4o-mini   & $0.732 \rightarrow 0.732$ & $+0.000$ [$-0.035$, $+0.027$] \\
gpt-5-mini    & $0.773 \rightarrow 0.774$ & $+0.001$ [$-0.014$, $+0.016$] \\
gpt-5.4       & $0.796 \rightarrow 0.794$ & $-0.002$ [$-0.014$, $+0.011$] \\
\midrule
\multicolumn{3}{@{}l}{\textit{JudgeBench}} \\
Gemma-4-31B   & $0.816 \rightarrow 0.844$ & $+0.029$ [$+0.017$, $+0.040$] \\
gpt-oss-120B  & $0.818 \rightarrow 0.836$ & $+0.018$ [$+0.003$, $+0.032$] \\
Qwen3.5-122B  & $0.868 \rightarrow 0.883$ & $+0.015$ [$+0.001$, $+0.030$] \\
Qwen3.5-27B   & $0.879 \rightarrow 0.899$ & $+0.021$ [$+0.007$, $+0.033$] \\
gpt-4o-mini   & $0.593 \rightarrow 0.574$ & $-0.018$ [$-0.093$, $+0.037$] \\
gpt-5-mini    & $1.000 \rightarrow 1.000$ & $+0.000$ [$+0.000$, $+0.000$] \\
gpt-5.4       & $0.867 \rightarrow 0.883$ & $+0.017$ [$-0.050$, $+0.083$] \\
\midrule
\multicolumn{3}{@{}l}{\textit{RewardBench}} \\
Gemma-4-31B   & $0.828 \rightarrow 0.859$ & $+0.031$ [$+0.021$, $+0.040$] \\
gpt-oss-120B  & $0.852 \rightarrow 0.885$ & $+0.033$ [$+0.022$, $+0.045$] \\
Qwen3.5-122B  & $0.835 \rightarrow 0.861$ & $+0.026$ [$+0.015$, $+0.037$] \\
Qwen3.5-27B   & $0.825 \rightarrow 0.851$ & $+0.025$ [$+0.014$, $+0.037$] \\
gpt-4o-mini   & $0.533 \rightarrow 0.567$ & $+0.033$ [$-0.033$, $+0.100$] \\
gpt-5-mini    & $0.650 \rightarrow 0.683$ & $+0.033$ [$+0.000$, $+0.083$] \\
gpt-5.4       & $0.667 \rightarrow 0.683$ & $+0.017$ [$+0.000$, $+0.050$] \\
\bottomrule
\end{tabular}%
}
\caption{
Accuracy measures agreement with the benchmark labels. We report the performance shift from a single evaluation pass (normal) to a two-direction average (avg). Ties are retained to ensure consistent denominators, as averaging inherently reduces tie rates. 95\% CIs are computed via question-level cluster bootstrapping (2,000 resamples), accounting for multiple comparisons per MT-Bench prompt.
}
\label{tab:decision}
\end{table}

\subsection{RQ4: Can Violations of Semantic Faithfulness Be Mitigated?}

Having established that all evaluated judges exhibit compromised semantic faithfulness, we investigate mitigating this deficiency during inference. To compare averaging methods at equal pass counts, we provide analyses with same-direction two-pass controls in Appendix~\ref{app:fourpass}.

Table~\ref{tab:remedy} demonstrates that bidirectional anchor averaging improves semantic robustness while maintaining human alignment. Relying on a single anchor direction arbitrarily restricts the leaderboard. For instance, $\tau_{\mathrm{dir}}$ is strictly less than 1 across all judges on SummEval, meaning the forward and reverse directions produce divergent rankings. We do not interpret $\tau_{\mathrm{dir}}$ independently as definitive proof of ranking distortion, as it is constrained by the inherent reproducibility of the judge. The floor-normalized ranking-level deficit detailed in Appendix~\ref{app:bootstrap} clarifies this effect. This distortion is substantial when the metric drops significantly, declining to 0.286 for gpt-4o-mini on BiGGen-Bench. On this benchmark, the top-performing system changes based on the anchor direction for two of the seven judges. Averaging scores from both directions reduces reliance on choosing a single scale direction.

We observe enhanced item-level agreement with human annotations in thirteen of fourteen configurations. The sole exception is gpt-5.4 on SummEval, which exhibits a negligible decrease of $0.002$. On BiGGen-Bench, improvements for Gemma-4-31B by $0.003$, gpt-5.4 by $0.014$, gpt-5-mini by $0.016$, and gpt-4o-mini by $0.040$ yield confidence intervals that include zero. The intervals for the remaining nine configurations strictly exclude zero, as further detailed in Appendix~\ref{app:bootstrap}. Crucially, these enhancements preserve system-level agreement because the confidence intervals for $\Delta\tau_{\mathrm{sys}}$ consistently encompass zero.

Table~\ref{tab:decision} evaluates bidirectional averaging on the three pairwise benchmarks. Relative to a single normal pass, accuracy increases in 17 of 21 judge--benchmark settings, remains unchanged in two, and decreases in two. Nine improvements have 95\% confidence intervals entirely above zero. All four open-weight judges improve on JudgeBench and RewardBench, with intervals excluding zero, whereas on MT-Bench only Gemma-4-31B has an interval entirely above zero. These results show that bidirectional averaging often improves agreement with benchmark labels, although the gains vary across judges and benchmarks.

\section{Conclusion}
We introduced \textit{semantic faithfulness}, a property that measures whether LLM judges follow the semantic meaning of evaluation rubrics rather than their numerical encoding. To probe this property, we proposed a simple scale-reversal protocol together with floor normalization, which contextualizes reversal sensitivity against intrinsic test--retest variability. Experiments on five meta-evaluation benchmarks show that all evaluated judges exhibit violations of semantic faithfulness, although the severity varies substantially across models. We further demonstrated that bidirectional anchor averaging effectively mitigates this failure mode while preserving agreement with human judgments. We hope our findings encourage future work to evaluate not only whether LLM judges agree with humans, but also whether they faithfully follow the rubrics they are given.

\section*{Limitations}
Our study has several limitations. First, we evaluate seven LLM judges: four open-weight models and three proprietary API models. Although we cover seven judges across open-weight and proprietary families, our conclusions remain limited to the evaluated model versions and API configurations. Second, our experimental design alters the rating scale length but restricts the scale type to integer ordinal scores with discrete semantic anchors. Our framework reverses this encoding and calibrates the resulting disagreement against the baseline floor of the judge. Although this methodology may extend to other semantic scoring schemes, further research is needed for continuous scores, binary verdicts, and native side-by-side protocols. In our current pairwise benchmarks, the judge still evaluates responses independently before deriving a final verdict.

Finally, bidirectional anchor averaging reduces reliance on choosing a single scale direction but doubles the inference cost by requiring two scoring passes per item. When this computational overhead is prohibitive, our floor-normalized audit provides a practical alternative to identify a reliable judge under a single direction. However, this audit merely indicates reliability without removing the inherent directional bias.

\section*{Ethical considerations}
This work analyzes the reliability of automatic evaluation systems and does not introduce new methods for generating harmful content. All experiments are conducted using publicly available benchmarks, open-weight models, and commercial APIs. We do not collect or annotate new human data, and no personally identifiable information is involved. We believe that improving the semantic faithfulness of LLM judges contributes to more transparent and trustworthy automatic evaluation practices.

\section*{Acknowledgement}
This work was supported by Korea Internet \& Security Agency(KISA) grant funded by the Korea government(PIPC) (No.RS-2026-25526342, Development of Technologies for Preventing Sensitive Information Inference and Risk Assessment in Foundation Model Operations). This work was also supported by the National Research Foundation of Korea(NRF) grant funded by the Korea government(MSIT) (RS-2026-25483747). In addition, this research was supported by Culture, Sports and Tourism R\&D Program through the Korea Creative Content Agency grant funded by the Ministry of Culture, Sports and Tourism in 2026 (Project Name: Develop AI agent technology to connect knowledge through public cultural facility-based discussion and communication, Project Number: RS-2026-25520645). Furthermore, this work was supported by Institute of Information \& communications Technology Planning \& Evaluation (IITP, AI Computing Support Project for R\&D) grant funded by the Korea government(MSIT) (High-Performance Research AI Computing Infrastructure Support at the 2 PFLOPS Scale, RS-2026-25505492). Lastly, we gratefully acknowledge VESSL for providing GPU resources.

\bibliography{custom}

@article{zheng2023judging,
  title={Judging llm-as-a-judge with mt-bench and chatbot arena},
  author={Zheng, Lianmin and Chiang, Wei-Lin and Sheng, Ying and Zhuang, Siyuan and Wu, Zhanghao and Zhuang, Yonghao and Lin, Zi and Li, Zhuohan and Li, Dacheng and Xing, Eric and others},
  journal={Advances in neural information processing systems},
  volume={36},
  pages={46595--46623},
  year={2023}
}

@inproceedings{lin-2004-rouge,
    title = "{ROUGE}: A Package for Automatic Evaluation of Summaries",
    author = "Lin, Chin-Yew",
    booktitle = "Text Summarization Branches Out",
    month = jul,
    year = "2004",
    address = "Barcelona, Spain",
    publisher = "Association for Computational Linguistics",
    url = "https://aclanthology.org/W04-1013/",
    pages = "74--81"
}

@article{zhang2019bertscore,
  title={Bertscore: Evaluating text generation with bert},
  author={Zhang, Tianyi and Kishore, Varsha and Wu, Felix and Weinberger, Kilian Q and Artzi, Yoav},
  journal={arXiv preprint arXiv:1904.09675},
  year={2019}
}

@inproceedings{sellam2020bleurt,
  title={BLEURT: Learning robust metrics for text generation},
  author={Sellam, Thibault and Das, Dipanjan and Parikh, Ankur},
  booktitle={Proceedings of the 58th annual meeting of the association for computational linguistics},
  pages={7881--7892},
  year={2020}
}

@article{fabbri2021summeval,
  title={Summeval: Re-evaluating summarization evaluation},
  author={Fabbri, Alexander R and Kry{\'s}ci{\'n}ski, Wojciech and McCann, Bryan and Xiong, Caiming and Socher, Richard and Radev, Dragomir},
  journal={Transactions of the Association for Computational Linguistics},
  volume={9},
  pages={391--409},
  year={2021},
  publisher={MIT Press One Rogers Street, Cambridge, MA 02142-1209, USA journals-info~…}
}

@inproceedings{deutsch2022limitations,
  title={On the limitations of reference-free evaluations of generated text},
  author={Deutsch, Daniel and Dror, Rotem and Roth, Dan},
  booktitle={Proceedings of the 2022 Conference on Empirical Methods in Natural Language Processing},
  pages={10960--10977},
  year={2022}
}

@article{kim2024biggen,
  title={The biggen bench: A principled benchmark for fine-grained evaluation of language models with language models},
  author={Kim, Seungone and Suk, Juyoung and Cho, Ji Yong and Longpre, Shayne and Kim, Chaeeun and Yoon, Dongkeun and Son, Guijin and Cho, Yejin and Shafayat, Sheikh and Baek, Jinheon and others},
  journal={arXiv preprint arXiv:2406.05761},
  year={2024}
}

@article{agarwal2025gpt,
  title={gpt-oss-120b \& gpt-oss-20b model card},
  author={Agarwal, Sandhini and Ahmad, Lama and Ai, Jason and Altman, Sam and Applebaum, Andy and Arbus, Edwin and Arora, Rahul K and Bai, Yu and Baker, Bowen and Bao, Haiming and others},
  journal={arXiv preprint arXiv:2508.10925},
  year={2025}
}

@article{team2026gemma,
  title={Gemma 4 technical report},
  author={Team, Gemma and Abd, Sherif El and Aggarwal, Vaibhav and Algayres, Robin and Andreev, Alek and Bachem, Olivier and Ballantyne, Ian and Brick, Cormac and C{\u{a}}rbune, Victor and Casbon, Michelle and others},
  journal={arXiv preprint arXiv:2607.02770},
  year={2026}
}

@inproceedings{kwon2023efficient,
  title={Efficient memory management for large language model serving with pagedattention},
  author={Kwon, Woosuk and Li, Zhuohan and Zhuang, Siyuan and Sheng, Ying and Zheng, Lianmin and Yu, Cody Hao and Gonzalez, Joseph and Zhang, Hao and Stoica, Ion},
  booktitle={Proceedings of the 29th symposium on operating systems principles},
  pages={611--626},
  year={2023}
}

@inproceedings{li2026evaluating,
  title={Evaluating scoring bias in llm-as-a-judge},
  author={Li, Qingquan and Dou, Shaoyu and Shao, Kailai and Chen, Chao and Hu, Haixiang},
  booktitle={International Conference on Database Systems for Advanced Applications},
  pages={19--34},
  year={2026},
  organization={Springer}
}

@inproceedings{lambert2025rewardbench,
  title={Rewardbench: Evaluating reward models for language modeling},
  author={Lambert, Nathan and Pyatkin, Valentina and Morrison, Jacob and Miranda, LJ and Lin, Bill Yuchen and Chandu, Khyathi and Dziri, Nouha and Kumar, Sachin and Zick, Tom and Choi, Yejin and others},
  booktitle={Findings of the Association for Computational Linguistics: NAACL 2025},
  pages={1755--1797},
  year={2025}
}

@article{tan2024judgebench,
  title={Judgebench: A benchmark for evaluating llm-based judges},
  author={Tan, Sijun and Zhuang, Siyuan and Montgomery, Kyle and Tang, William Y and Cuadron, Alejandro and Wang, Chenguang and Popa, Raluca Ada and Stoica, Ion},
  journal={arXiv preprint arXiv:2410.12784},
  year={2024}
}

@article{hurst2024gpt,
  title={Gpt-4o system card},
  author={Hurst, Aaron and Lerer, Adam and Goucher, Adam P and Perelman, Adam and Ramesh, Aditya and Clark, Aidan and Ostrow, AJ and Welihinda, Akila and Hayes, Alan and Radford, Alec and others},
  journal={arXiv preprint arXiv:2410.21276},
  year={2024}
}

@article{singh2025openai,
  title={Openai gpt-5 system card},
  author={Singh, Aaditya and Fry, Adam and Perelman, Adam and Tart, Adam and Ganesh, Adi and El-Kishky, Ahmed and McLaughlin, Aidan and Low, Aiden and Ostrow, AJ and Ananthram, Akhila and others},
  journal={arXiv preprint arXiv:2601.03267},
  year={2025}
}

@inproceedings{ye2025justice,
  title={Justice or prejudice? quantifying biases in llm-as-a-judge},
  author={Ye, Jiayi and Wang, Yanbo and Huang, Yue and Chen, Dongping and Zhang, Qihui and Moniz, Nuno and Gao, Tian and Geyer, Werner and Huang, Chao and Chen, Pin-Yu and others},
  booktitle={International Conference on Learning Representations},
  volume={2025},
  pages={102351--102390},
  year={2025}
}

@inproceedings{chen2024humans,
  title={Humans or LLMs as the judge? a study on judgement bias},
  author={Chen, Guiming Hardy and Chen, Shunian and Liu, Ziche and Jiang, Feng and Wang, Benyou},
  booktitle={Proceedings of the 2024 Conference on Empirical Methods in Natural Language Processing},
  pages={8301--8327},
  year={2024}
}

@article{gu2024survey,
  title={A survey on llm-as-a-judge},
  author={Gu, Jiawei and Jiang, Xuhui and Shi, Zhichao and Tan, Hexiang and Zhai, Xuehao and Xu, Chengjin and Li, Wei and Shen, Yinghan and Ma, Shengjie and Liu, Honghao and others},
  journal={The Innovation},
  year={2024},
  publisher={Elsevier}
}

@inproceedings{liu2024calibrating,
  title={Calibrating llm-based evaluator},
  author={Liu, Yuxuan and Yang, Tianchi and Huang, Shaohan and Zhang, Zihan and Huang, Haizhen and Wei, Furu and Deng, Weiwei and Sun, Feng and Zhang, Qi},
  booktitle={Proceedings of the 2024 joint international conference on computational linguistics, language resources and evaluation (lrec-coling 2024)},
  pages={2638--2656},
  year={2024}
}

@article{stureborg2024large,
  title={Large language models are inconsistent and biased evaluators},
  author={Stureborg, Rickard and Alikaniotis, Dimitris and Suhara, Yoshi},
  journal={arXiv preprint arXiv:2405.01724},
  year={2024}
}

@article{haldar2025rating,
  title={Rating roulette: Self-inconsistency in LLM-as-a-judge frameworks},
  author={Haldar, Rajarshi and Hockenmaier, Julia},
  journal={arXiv preprint arXiv:2510.27106},
  year={2025}
}

@inproceedings{liu2023g,
  title={G-eval: NLG evaluation using gpt-4 with better human alignment},
  author={Liu, Yang and Iter, Dan and Xu, Yichong and Wang, Shuohang and Xu, Ruochen and Zhu, Chenguang},
  booktitle={Proceedings of the 2023 conference on empirical methods in natural language processing},
  pages={2511--2522},
  year={2023}
}

@inproceedings{thakur2025judging,
  title={Judging the judges: Evaluating alignment and vulnerabilities in llms-as-judges},
  author={Thakur, Aman Singh and Choudhary, Kartik and Ramayapally, Venkat Srinik and Vaidyanathan, Sankaran and Hupkes, Dieuwke},
  booktitle={Proceedings of the Fourth Workshop on Generation, Evaluation and Metrics (GEM$^2$)},
  pages={404--430},
  year={2025}
}

@inproceedings{mathur2020tangled,
  title={Tangled up in BLEU: Reevaluating the evaluation of automatic machine translation evaluation metrics},
  author={Mathur, Nitika and Baldwin, Timothy and Cohn, Trevor},
  booktitle={Proceedings of the 58th annual meeting of the association for computational linguistics},
  pages={4984--4997},
  year={2020}
}

@inproceedings{deutsch2022re,
  title={Re-examining system-level correlations of automatic summarization evaluation metrics},
  author={Deutsch, Daniel and Dror, Rotem and Roth, Dan},
  booktitle={Proceedings of the 2022 Conference of the North American Chapter of the Association for Computational Linguistics: Human Language Technologies},
  pages={6038--6052},
  year={2022}
}

@article{deutsch2021statistical,
  title={A statistical analysis of summarization evaluation metrics using resampling methods},
  author={Deutsch, Daniel and Dror, Rotem and Roth, Dan},
  journal={Transactions of the Association for Computational Linguistics},
  volume={9},
  pages={1132--1146},
  year={2021},
  publisher={MIT Press One Rogers Street, Cambridge, MA 02142-1209, USA journals-info~…}
}

@article{asato2026mitigating,
  title={Mitigating Scoring Bias in LLM-as-a-Judge via Random Number Generation},
  author={Asato, Yuma and Shirai, Kiyoaki and Kertkeidkachorn, Natthawut},
  journal={arXiv preprint arXiv:2608.05726},
  year={2026}
}

@Article{math11041006,
AUTHOR = {Lee, Seungjun and Lee, Jungseob and Moon, Hyeonseok and Park, Chanjun and Seo, Jaehyung and Eo, Sugyeong and Koo, Seonmin and Lim, Heuiseok},
TITLE = {A Survey on Evaluation Metrics for Machine Translation},
JOURNAL = {Mathematics},
VOLUME = {11},
YEAR = {2023},
NUMBER = {4},
ARTICLE-NUMBER = {1006},
URL = {https://www.mdpi.com/2227-7390/11/4/1006},
ISSN = {2227-7390},
DOI = {10.3390/math11041006}
}

\clearpage
\newpage
\appendix
\section{Consistency Threshold \texorpdfstring{$\varepsilon$}{epsilon}}
\label{app:epsilon}

We formulate the item-level consistency deficit as $\Delta_{\mathrm{exact}} = A_{\mathrm{floor}} - A_{\mathrm{flip}}$, which is evaluated relative to a tolerance margin of $\varepsilon = 0.02$. This margin accounts for the inherent decoding noise expected across independent evaluation passes under identical rubrics. To provide a rigorous evaluation, we report the continuous deficit with bootstrap confidence intervals. We additionally mark 0.02 as a practical-effect reference rather than a statistical decision threshold. This contextualizes substantive systematic bias against the natural variance of an evaluator.

Calibrating against floor performance is essential, as absolute agreement metrics are uninformative in isolation. For instance, an absolute flip agreement of $A_{\mathrm{flip}} = 0.75$ indicates substantial bias for a highly reliable evaluator ($A_{\mathrm{floor}} = 0.98$), given that the $0.23$ deficit vastly exceeds the $\varepsilon$ reference. Conversely, the identical $A_{\mathrm{flip}} = 0.75$ for a less consistent evaluator ($A_{\mathrm{floor}} = 0.75$) yields a negligible deficit that falls within expected variance. Without accounting for an evaluator's intrinsic self-consistency, identical flip agreements could lead to contradictory conclusions regarding bias. By framing the deficit relatively, our formulation accurately captures the critical performance disparities observed between the high-floor and low-floor judges in our experiments.

\section{Four-Pass Repeatability and Matched Averaging Comparisons}

\label{app:fourpass}

\subsection{Conditions and comparison sets}

\label{app:fourpass-conditions}

We use four recorded scoring passes: $N_1$ (\texttt{normal}), $N_2$ (\texttt{normal2}), $R_1$ (\texttt{reversed}), and $R_2$ (\texttt{reversed2}). The two normal passes use the same normal rubric, and the two reversed passes use the same reversed rubric. Reversed raw scores are restored to the original quality scale as $\widetilde{R}_k=(s_{\min}+s_{\max})-R_k$. The constant is 6 on the five-point benchmarks and 11 on MT-Bench.

We report two analyses with separate complete-case sets: a reversed-repeat diagnostic for GPT-4o-mini on SummEval, and a utility comparison using recorded scores for four open-weight judges across five benchmarks. Each analysis holds the evaluated observations fixed across its compared conditions. The latter reuses recorded scores without additional model calls and does not require a common subset across judges; it estimates within-judge method contrasts rather than cross-judge rankings.

\subsection{Reversed-repeat baseline}

\label{app:reverse-repeat}

Let $A_N=A(N_1,N_2)$, $A_R=A(\widetilde{R}_1,\widetilde{R}_2)$, and $A_{\mathrm{flip}}=A(N_1,\widetilde{R}_1)$, where $A$ is exact-score agreement. On the same four-pass observations, Equation~\ref{eq:repeat_baseline_deficits} defines the normal-baseline and symmetric-baseline deficits:

\begin{equation}
\begin{aligned}
\Delta_N &= A_N-A_{\mathrm{flip}},\\
\Delta_{\mathrm{sym}} &= \frac{A_N+A_R}{2}-A_{\mathrm{flip}}.
\end{aligned}
\label{eq:repeat_baseline_deficits}
\end{equation}

The GPT-4o-mini check retains 1,413 response--criterion observations from 30 SummEval source articles across coherence, consistency, and relevance. All three agreements use the four-pass intersection. We use 2,000 source-article cluster bootstrap resamples for 95\% intervals, preserving responses and criteria within an article.

\begin{table}[t]
\centering\small
\begin{tabular*}{\columnwidth}{@{\extracolsep{\fill}}lr@{}}
\toprule
Measure & Estimate [95\% CI] \\
\midrule
$A_N$ & +90.23 [+88.30, +92.06] \\
$A_R$ & +86.62 [+84.73, +88.49] \\
$A_{\mathrm{flip}}$ & +65.04 [+61.42, +68.72] \\
$\Delta_N$ & +25.19 [+21.26, +29.21] \\
$\Delta_{\mathrm{sym}}$ & +23.39 [+19.44, +27.45] \\
\bottomrule
\end{tabular*}
\caption{Direction-specific repeatability and reversal deficits for GPT-4o-mini on the shared four-pass SummEval subset. Agreement is expressed in percent and deficits in percentage points. This model-specific subset is separate from the seven-judge common-sample analysis and from the open-weight utility analysis below.}
\label{tab:reverse-repeat}
\end{table}

Reversed-repeat agreement is lower than normal-repeat agreement. Averaging the two repeat baselines reduces the deficit from 25.19 to 23.39 pp, while cross-direction agreement remains below either repeat agreement. This supports excess cross-direction disagreement relative to both baselines on this subset.

\begin{table*}[t]
\centering\small
\setlength{\tabcolsep}{5pt}
\begin{tabular}{@{}llrr@{}}
\toprule
Benchmark & Judge & NR-NN [95\% CI] & NR-RR [95\% CI] \\
\midrule
SummEval & Gemma-4-31B & $-0.0065 [-0.0113, -0.0019]$ & $+0.0010 [-0.0037, +0.0059]$ \\
SummEval & gpt-oss-120B & $-0.0003 [-0.0081, +0.0079]$ & $+0.0058 [-0.0013, +0.0128]$ \\
SummEval & Qwen3.5-122B & $-0.0007 [-0.0064, +0.0051]$ & $+0.0010 [-0.0067, +0.0085]$ \\
SummEval & Qwen3.5-27B & $+0.0034 [-0.0029, +0.0099]$ & $+0.0009 [-0.0050, +0.0067]$ \\
BiGGen-Bench & Gemma-4-31B & $-0.0061 [-0.0109, -0.0011]$ & $+0.0025 [-0.0030, +0.0080]$ \\
BiGGen-Bench & gpt-oss-120B & $-0.0011 [-0.0070, +0.0049]$ & $-0.0051 [-0.0110, +0.0010]$ \\
BiGGen-Bench & Qwen3.5-122B & $-0.0000 [-0.0065, +0.0065]$ & $-0.0009 [-0.0071, +0.0051]$ \\
BiGGen-Bench & Qwen3.5-27B & $-0.0034 [-0.0091, +0.0021]$ & $-0.0027 [-0.0087, +0.0034]$ \\
MT-Bench & Gemma-4-31B & $+0.52 [+0.04, +1.02]$ & $+0.00 [-0.69, +0.68]$ \\
MT-Bench & gpt-oss-120B & $+0.41 [-0.42, +1.23]$ & $+0.51 [-0.34, +1.44]$ \\
MT-Bench & Qwen3.5-122B & $-0.18 [-1.15, +0.75]$ & $+0.38 [-0.57, +1.32]$ \\
MT-Bench & Qwen3.5-27B & $-0.16 [-0.97, +0.65]$ & $+0.59 [-0.18, +1.41]$ \\
JudgeBench & Gemma-4-31B & $+1.56 [+0.59, +2.60]$ & $+0.68 [-0.50, +1.96]$ \\
JudgeBench & gpt-oss-120B & $+0.40 [-1.23, +2.03]$ & $+0.24 [-1.15, +1.62]$ \\
JudgeBench & Qwen3.5-122B & $-1.26 [-2.67, +0.12]$ & $-1.15 [-2.55, +0.12]$ \\
JudgeBench & Qwen3.5-27B & $+0.91 [-0.27, +2.11]$ & $+0.45 [-0.72, +1.63]$ \\
RewardBench & Gemma-4-31B & $+2.84 [+1.90, +3.79]$ & $+0.19 [-0.51, +0.90]$ \\
RewardBench & gpt-oss-120B & $+2.12 [+1.01, +3.30]$ & $-0.07 [-1.13, +1.01]$ \\
RewardBench & Qwen3.5-122B & $+0.29 [-0.73, +1.31]$ & $+0.94 [-0.07, +1.98]$ \\
RewardBench & Qwen3.5-27B & $+0.21 [-0.99, +1.42]$ & $-0.49 [-1.55, +0.56]$ \\
\bottomrule
\end{tabular}
\caption{Direct utility contrasts on judge-specific four-pass complete cases. Positive differences favor NR. Accuracy differences are percentage points; graded-rating contrasts use standard Kendall $\tau_b$. Comparisons with other correlation tables must account for differences in sample selection and tie handling.}
\label{tab:paired-utility}
\end{table*}

\subsection{Utility beyond same-direction repetition}

\label{app:paired-utility}

To assess the utility of mixing directions beyond repeated averaging, we compare three two-pass scoring rules summarized in Table~\ref{tab:scoring-rules}.

\begin{table}[H]
\centering\small
\begin{tabular*}{\columnwidth}{@{\extracolsep{\fill}}llc@{}}
\toprule
Rule & Passes & Averaged score \\
\midrule
NN & Two normal & $(N_1+N_2)/2$ \\
NR & Mixed & $(N_1+\widetilde{R}_1)/2$ \\
RR & Two reversed & $(\widetilde{R}_1+\widetilde{R}_2)/2$ \\
\bottomrule
\end{tabular*}
\caption{Two-pass scoring rules. Reversed scores are remapped before averaging.}
\label{tab:scoring-rules}
\end{table}

We retain half-integer averages without rounding. Each rule uses two judge calls per response; four passes supply the data for the comparison. For each judge and benchmark, all rules use the intersection of valid scores in all four passes and available external labels. Missing, malformed, or out-of-range scores are excluded jointly. For pairwise benchmarks, both responses must have valid scores in every pass. Human ties or unspecified winners are excluded, whereas judge score ties receive accuracy credit of 0.5.

SummEval and BiGGen-Bench use standard Kendall $\tau_b$ against human ratings. SummEval averages criterion-level $\tau_b$ equally over coherence, consistency, and relevance. The three pairwise benchmarks use tie-adjusted accuracy.

The additional utility of mixing directions is setting-dependent. For Gemma-4-31B on RewardBench, NR exceeds NN by 2.84 pp [1.90, 3.79], but its advantage over RR is only 0.19 pp [-0.51, 0.90]. On SummEval, the same judge has lower $\tau_b$ under NR than NN (-0.0065 [-0.0113, -0.0019]). Across the 20 settings, no setting has both direct-contrast intervals entirely above zero. This is a descriptive count, not a global test or evidence of equivalence. Swapping the two inputs of an average leaves the result unchanged, but this symmetry does not establish cancellation of judgment errors, restored judge invariance, or preserved human alignment.

Averaging can reduce noise variance, but the amount depends on the variances and covariance of the two errors. Halving the variance requires uncorrelated errors of equal variance. Complete cancellation of directional bias requires biases of equal magnitude and opposite sign after remapping. Scale reversal alone does not establish these conditions.

The comparison matches pass count and observations, not direction-specific noise, score distributions, or tie frequencies. Averaging integer scores permits half-integer values and can change tie frequencies, which may affect agreement metrics. Complete-case filtering can introduce selection effects, and the results are conditional on the recorded runs. The reversed-repeat check and utility comparisons therefore support descriptive assessments of baseline sensitivity and averaging utility, without a causal decomposition of scale-direction effects.

\section{Judge Prompt Templates}
\label{app:prompts}

Each judge call fills the placeholders in braces with the corresponding
fields of the evaluation instance.
The \texttt{normal} and \texttt{normal2} conditions use the identical prompt,
while only the reversed condition changes the score direction.
Apart from the score scale, the prompts are character-for-character identical.

\subsection{SummEval}
\begin{PromptBlock}{Prompt Template: SummEval}
You are a strict, expert judge evaluating the quality of document summarization.
Based on the [Source Document], [Summarization Instruction],
[Reference Summaries], and the [Summary] produced by the target system
provided below, you must evaluate ONLY the single specified
[Evaluation Criterion].

# Input Data

[Source Document]
{document}

[Summarization Instruction]
{question}

[Reference Summaries]
{golden_answer}

[Summary]
{model_response}

[Evaluation Criterion]
{rubric}

(Note: this criterion is scored on a 5-point scale from
1 (very poor) to 5 (excellent).)

# Evaluation Guidelines (must be followed)

1. Evaluate strictly according to the definition of the single
   [Evaluation Criterion] above, and do not consider any other aspect.

   - consistency (factual alignment): judge hallucination/distortion ONLY
     by checking against the [Source Document].

   - coherence (structural organization) and fluency (sentence quality):
     judge the [Summary] on its own, independent of the
     [Source Document] and [Reference Summaries].

   - relevance (important content): judge whether the [Summary]
     captures the important content of the [Source Document]
     while excluding unnecessary or redundant information.

2. The [Reference Summaries] are only a guide for gauging what content
   is important. Even if the [Summary] differs from them in wording or
   structure, do NOT deduct points for a valid alternative paraphrase.
   There is no single correct answer.

3. Chain of Thought: before assigning the final score,
   you MUST first write concrete reasoning that analyzes the summary
   against the single evaluation criterion above and derives the score.

4. Ensuring discrimination: assign 5 (top score) ONLY when it is hard
   to find any flaw on this criterion.
   If there is even a minor flaw, score 4 or below.

# Output Format

Output the result ONLY in the JSON format below.
Do not include any other explanation.

For criterion_name, use the exact key of the given evaluation criterion
(e.g., rubric_coherence).

{
  "criterion_name": "[the key of the evaluation criterion]",
  "reasoning": "[analysis of the summary according to the definition of the criterion]",
  "score": [an integer between 1 and 5]
}
\end{PromptBlock}

\subsection{BiGGen-Bench}
\begin{PromptBlock}{Prompt Template: BiGGen-Bench}
You are a fair, strict expert judge assessing the quality of a language model's response, following the absolute (direct-assessment) grading protocol of BiGGen-Bench.
You are given an instruction (it may contain an input), the response to evaluate, a reference answer that is an example of a top-quality (Score 5) response, and an instance-specific score rubric that gives a description for each score from 1 to 5. Evaluate the response ONLY against the given rubric.

# Input Data
[Instruction to evaluate]
{question}

[Response to evaluate]
{model_response}

[Reference Answer (Score 5)]
{golden_answer}

[Score Rubric]
{rubric}
(Note: this is scored on a 5-point scale from 1 (worst, matching the Score 1 description) to 5 (best, matching the Score 5 description). The reference answer above is an example of a Score 5 response.)

# Evaluation Guidelines (must be followed)
1. Assess the response strictly against the [Score Rubric] above, not by any general impression. Match the response to the score description (1-5) it best fits.
2. The [Reference Answer] is only a Score-5 exemplar to calibrate the top of the scale; a valid alternative that satisfies the rubric is not penalized merely for differing from it in wording.
3. Chain of Thought: before assigning the final score, you MUST first write concrete reasoning that compares the response to the rubric's score descriptions and derives the score.
4. Ensuring discrimination: assign 5 (top score) ONLY when the response clearly matches the Score 5 description with no notable flaw; otherwise score 4 or below.

# Output Format
Output the result ONLY in the JSON format below. Do not include any other explanation.
For criterion_name, use the exact key of the given rubric (i.e., rubric_quality).

{
  "criterion_name": "rubric_quality",
  "reasoning": "[comparison of the response against the rubric's score descriptions]",
  "score": [an integer between 1 and 5]
}
\end{PromptBlock}

\subsection{MT-Bench}
\begin{PromptBlock}{Prompt Template: MT-Bench}
You are a strict, expert judge assessing a single answer produced by an AI assistant, following the single-answer grading protocol of MT-Bench.
Based on the [Exchange] and the [Answer] provided below, you must evaluate ONLY the single specified [Evaluation Criterion]. You are shown one answer at a time; no competing answer is available, so judge this answer on its own merits.

# Input Data
[Exchange]
{question}

[Answer]
{model_response}

[Evaluation Criterion]
{rubric}
(Note: this criterion is scored on a 10-point scale from 1 (very poor) to 10 (excellent).)

# Evaluation Guidelines (must be followed)
1. Evaluate strictly according to the definition of the single [Evaluation Criterion] above, and do not consider any other aspect.
2. If the [Exchange] contains an earlier turn, judge ONLY the [Answer] to the final user question, but judge it in the context of that earlier turn: an answer that ignores the constraint carried over from the first turn is a poor answer. Do not score the earlier turn's answer.
3. Chain of Thought: before assigning the final score, you MUST first write concrete reasoning that analyzes the answer against the single evaluation criterion above and derives the score.
4. Ensuring discrimination: assign 10 (top score) ONLY when it is hard to find any flaw on this criterion. If there is even a minor flaw, score 9 or below.

# Output Format
Output the result ONLY in the JSON format below. Do not include any other explanation.
For criterion_name, use the exact key of the given evaluation criterion (i.e., rubric_quality).

{
  "criterion_name": "rubric_quality",
  "reasoning": "[analysis of the answer according to the definition of the criterion]",
  "score": [an integer between 1 and 10]
}
\end{PromptBlock}

\subsection{JudgeBench}
\begin{PromptBlock}{Prompt Template: JudgeBench}
You are a strict, expert judge assessing a single response produced by an AI assistant.
Based on the [Question] and the [Response] provided below, you must evaluate ONLY the single specified [Evaluation Criterion]. You are shown one response at a time; no competing response is available, so judge this response on its own merits.

# Input Data
[Question]
{question}

[Response]
{model_response}

[Evaluation Criterion]
{rubric}
(Note: this criterion is scored on a 5-point scale from 1 (very poor) to 5 (excellent).)

# Evaluation Guidelines (must be followed)
1. Evaluate strictly according to the definition of the single [Evaluation Criterion] above, and do not consider any other aspect.
2. Work the problem yourself first: derive the correct answer to the [Question] independently, then check whether the [Response] reaches it and whether every step it takes is valid. A confident, well-formatted, or lengthy response that lands on a wrong answer is a low-scoring response.
3. Chain of Thought: before assigning the final score, you MUST first write concrete reasoning that analyzes the response against the single evaluation criterion above and derives the score.
4. Ensuring discrimination: assign 5 (top score) ONLY when it is hard to find any flaw on this criterion. If there is even a minor flaw, score 4 or below.

# Output Format
Output the result ONLY in the JSON format below. Do not include any other explanation.
For criterion_name, use the exact key of the given evaluation criterion (i.e., rubric_correctness).

{
  "criterion_name": "rubric_correctness",
  "reasoning": "[analysis of the response according to the definition of the criterion]",
  "score": [an integer between 1 and 5]
}
```

The `{rubric}` placeholder is filled with the following criterion, which is the same for every JudgeBench item and every pass.

```
- "rubric_correctness": Factual and logical correctness of the response. Judge whether the response reaches the correct final answer and whether every step of its reasoning is sound. A response that is fluent, well-structured, confident, or longer must still be scored low if its final answer is wrong or any reasoning step is invalid. Presentation, style, and length are irrelevant to this criterion.
\end{PromptBlock}

\subsection{RewardBench}
\begin{PromptBlock}{Prompt Template: RewardBench}
You are a strict, expert judge assessing a single response produced by an AI assistant.
Based on the [User Request] and the [Response] provided below, you must evaluate ONLY the single specified [Evaluation Criterion]. You are shown one response at a time; no competing response is available, so judge this response on its own merits.

# Input Data
[User Request]
{question}

[Response]
{model_response}

[Evaluation Criterion]
{rubric}
(Note: this criterion is scored on a 5-point scale from 1 (very poor) to 5 (excellent).)

# Evaluation Guidelines (must be followed)
1. Evaluate strictly according to the definition of the single [Evaluation Criterion] above, and do not consider any other aspect.
2. Weigh helpfulness and safety together as one judgment. First decide whether the [User Request] is genuinely harmful: if it is, a response that refuses or safely redirects is good and one that complies is poor; if it is benign, a response that refuses, hedges excessively, or lectures the user is poor no matter how polite it is.
3. Chain of Thought: before assigning the final score, you MUST first write concrete reasoning that analyzes the response against the single evaluation criterion above and derives the score.
4. Ensuring discrimination: assign 5 (top score) ONLY when it is hard to find any flaw on this criterion. If there is even a minor flaw, score 4 or below.

# Output Format
Output the result ONLY in the JSON format below. Do not include any other explanation.
For criterion_name, use the exact key of the given evaluation criterion (i.e., rubric_quality).

{
  "criterion_name": "rubric_quality",
  "reasoning": "[analysis of the response according to the definition of the criterion]",
  "score": [an integer between 1 and 5]
}
```

The `{rubric}` placeholder is filled with the following criterion, which is the same for every RewardBench item and every pass.

```
- "rubric_quality": Overall quality of the response as an answer to the user's request, combining helpfulness and safety into a single judgment. A high-quality response follows the instruction accurately, is factually correct and genuinely useful, and is appropriately safe: it refuses or safely redirects a genuinely harmful request, while it does NOT refuse, hedge, or lecture in response to a benign request. Length, confident tone, and elaborate formatting do not by themselves make a response better.
\end{PromptBlock}

\section{Bias Deficits and Human Agreement}
\label{app:bootstrap}
Table~\ref{tab:deficits} reports item- and system-level reversal deficits as point estimates. The ranking-level deficit is defined as $\Delta_{\mathrm{rank}} = \tau(\bar{s}^{\mathrm{n}}, \bar{s}^{\mathrm{n2}}) - \tau_{\mathrm{dir}}$, where $\tau$ is Kendall's $\tau_b$, $\bar{s}^{\mathrm{n}}$ and $\bar{s}^{\mathrm{n2}}$ are system-level mean scores from two identical normal passes, and $\tau_{\mathrm{dir}}$ is the consistency between the normal and reversed passes. Table~\ref{tab:remedy-ci} reports changes in human agreement ($\Delta\tau_{\mathrm{item}}$ and $\Delta\tau_{\mathrm{sys}}$) with confidence intervals.

We estimate the intervals in Tables~\ref{tab:main} and~\ref{tab:remedy-ci} using a cluster bootstrap ($B = 2000$). By resampling natural clusters such as source articles, instances, or response pairs rather than individual judge calls, we preserve within-cluster correlations and accurately capture population-level variance. Note that the evaluation sets vary across analyses to fit their specific requirements: Tables~\ref{tab:main} and~\ref{tab:deficits} require a three-way intersection of normal, reversed, and normal2 passes to establish the baseline floor, whereas Table~\ref{tab:remedy-ci} intersects only normal and reversed passes with human ratings. Finally, Table~\ref{tab:decision} resamples questions rather than individual comparisons to handle prompt-level dependencies in pairwise data. Consequently, these tables are complementary and their cell counts are not directly comparable.

\begin{table}[!htbp]
\centering
\small
\renewcommand{\arraystretch}{0.9}
\setlength{\tabcolsep}{6pt}
\begin{tabular*}{\columnwidth}{@{\extracolsep{\fill}}lcc@{}}
\toprule
Judge & $\Delta_{\mathrm{exact}}$ & $\Delta_{\mathrm{rank}}$ \\
\midrule
\multicolumn{3}{@{}l}{\textit{SummEval}} \\
Gemma-4-31B   & $+0.169$ & $+0.022$ \\
gpt-oss-120b  & $+0.077$ & $+0.004$ \\
Qwen3.5-122B  & $+0.049$ & $-0.010$ \\
Qwen3.5-27B   & $+0.027$ & $-0.003$ \\
gpt-4o-mini   & $+0.249$ & $+0.085$ \\
gpt-5-mini    & $+0.042$ & $+0.029$ \\
gpt-5.4       & $+0.042$ & $-0.016$ \\
\midrule
\multicolumn{3}{@{}l}{\textit{BiGGen-Bench}} \\
Gemma-4-31B   & $+0.123$ & $\phantom{-}0.000$ \\
gpt-oss-120b  & $+0.044$ & $+0.333$ \\
Qwen3.5-122B  & $+0.064$ & $\phantom{-}0.000$ \\
Qwen3.5-27B   & $+0.053$ & $\phantom{-}0.000$ \\
gpt-4o-mini   & $+0.242$ & $+0.486$ \\
gpt-5-mini    & $+0.008$ & $\phantom{-}0.000$ \\
gpt-5.4       & $+0.042$ & $\phantom{-}0.000$ \\
\midrule
\multicolumn{3}{@{}l}{\textit{MT-Bench}} \\
Gemma-4-31B   & $+0.333$ & $\phantom{-}0.000$ \\
gpt-oss-120b  & $+0.124$ & $-0.133$ \\
Qwen3.5-122B  & $+0.150$ & $\phantom{-}0.000$ \\
Qwen3.5-27B   & $+0.120$ & $+0.133$ \\
gpt-4o-mini   & $+0.272$ & $+0.133$ \\
gpt-5-mini    & $+0.067$ & $\phantom{-}0.000$ \\
gpt-5.4       & $+0.155$ & $\phantom{-}0.000$ \\
\midrule
\multicolumn{3}{@{}l}{\textit{JudgeBench}} \\
Gemma-4-31B   & $+0.140$ & $\phantom{-}0.000$ \\
gpt-oss-120b  & $+0.022$ & $\phantom{-}0.000$ \\
Qwen3.5-122B  & $+0.011$ & $\phantom{-}0.000$ \\
Qwen3.5-27B   & $+0.044$ & $\phantom{-}0.000$ \\
gpt-4o-mini   & $+0.161$ & $+2.000$ \\
gpt-5-mini    & $+0.017$ & $\phantom{-}0.000$ \\
gpt-5.4       & $\phantom{-}0.000$ & $\phantom{-}0.000$ \\
\midrule
\multicolumn{3}{@{}l}{\textit{RewardBench}} \\
Gemma-4-31B   & $+0.222$ & $\phantom{-}0.000$ \\
gpt-oss-120b  & $+0.082$ & $+2.000$ \\
Qwen3.5-122B  & $+0.063$ & $\phantom{-}0.000$ \\
Qwen3.5-27B   & $+0.045$ & $\phantom{-}0.000$ \\
gpt-4o-mini   & $+0.067$ & $\phantom{-}0.000$ \\
gpt-5-mini    & $+0.117$ & $\phantom{-}0.000$ \\
gpt-5.4       & $+0.017$ & $\phantom{-}0.000$ \\
\bottomrule
\end{tabular*}
\caption{Item- and system-level reversal deficits ($\Delta_{\mathrm{exact}}$ and $\Delta_{\mathrm{rank}}$), reported as point estimates.}
\label{tab:deficits}
\end{table}

\begin{table}[!htbp]
\centering
\small
\setlength{\tabcolsep}{6pt}
\resizebox{\columnwidth}{!}{%
\begin{tabular}{@{}lcc@{}}
\toprule
Judge & $\Delta\tau_{\mathrm{item}}$ [95\% CI] & $\Delta\tau_{\mathrm{sys}}$ [95\% CI] \\
\midrule
\multicolumn{3}{@{}l}{\textit{SummEval}} \\
Gemma-4-31B   & $+0.015$ [$+0.010$, $+0.019$] & $-0.011$ [$-0.035$, $+0.020$] \\
gpt-oss-120b  & $+0.019$ [$+0.012$, $+0.027$] & $+0.030$ [$-0.015$, $+0.054$] \\
Qwen3.5-122B  & $+0.021$ [$+0.016$, $+0.026$] & $\phantom{-}0.000$ [$-0.030$, $+0.033$] \\
Qwen3.5-27B   & $+0.019$ [$+0.014$, $+0.025$] & $+0.015$ [$-0.026$, $+0.044$] \\
gpt-4o-mini   & $+0.010$ [$+0.001$, $+0.020$] & $-0.012$ [$-0.046$, $+0.059$] \\
gpt-5-mini    & $+0.018$ [$+0.011$, $+0.025$] & $-0.013$ [$-0.030$, $+0.045$] \\
gpt-5.4       & $-0.002$ [$-0.012$, $+0.010$] & $-0.026$ [$-0.042$, $+0.046$] \\
\midrule
\multicolumn{3}{@{}l}{\textit{BiGGen-Bench}} \\
Gemma-4-31B   & $+0.003$ [$-0.001$, $+0.007$] & $\phantom{-}0.000^{\dagger}$ \\
gpt-oss-120b  & $+0.013$ [$+0.008$, $+0.019$] & $\phantom{-}0.000^{\dagger}$ \\
Qwen3.5-122B  & $+0.013$ [$+0.009$, $+0.018$] & $\phantom{-}0.000^{\dagger}$ \\
Qwen3.5-27B   & $+0.009$ [$+0.005$, $+0.013$] & $\phantom{-}0.000^{\dagger}$ \\
gpt-4o-mini   & $+0.040$ [$-0.007$, $+0.091$] & $+0.204^{\dagger}$ \\
gpt-5-mini    & $+0.016$ [$-0.011$, $+0.048$] & $\phantom{-}0.000^{\dagger}$ \\
gpt-5.4       & $+0.014$ [$-0.010$, $+0.041$] & $\phantom{-}0.000^{\dagger}$ \\
\bottomrule
\end{tabular}%
}
\caption{Changes in item- and system-level Kendall correlation with human ratings after bidirectional averaging, relative to a single normal pass. Brackets: 95\% cluster bootstrap CIs (2,000 resamples). $\dagger$: point estimate only for BiGGen-Bench, which has four systems.}
\label{tab:remedy-ci}
\end{table}

\end{document}